\documentclass[runningheads]{llncs}

\usepackage{eccv}

\usepackage{eccvabbrv}

\usepackage{graphicx}
\usepackage{booktabs}

\usepackage[accsupp]{axessibility}  

\usepackage[pagebackref,breaklinks,colorlinks,citecolor=eccvblue]{hyperref}

\usepackage{orcidlink}

\usepackage{xcolor}
\usepackage{colortbl}
\definecolor{gray0}{gray}{0.9}

\usepackage{multirow}
\usepackage{bbding}
\usepackage{pifont}
\usepackage{utfsym}
\newcommand{\cmark}{\ding{51}\xspace}%
\definecolor{raycolor}{RGB}{255,192,0}

\usepackage{floatrow}
\newfloatcommand{capbtabbox}{table}[][\FBwidth]

\begin{document}

\title{Ray Denoising: Depth-aware Hard Negative Sampling for Multi-view 3D Object Detection}

\titlerunning{Abbreviated paper title}

\author{Feng Liu\inst{1}\thanks{Work was done during internship.}\quad 
Tengteng Huang\inst{2}\quad
Qianjing Zhang\inst{2}\quad
Haotian Yao\inst{2}\quad
Chi Zhang\inst{2}\\
Fang Wan\inst{1}\quad
Qixiang Ye\inst{1}\quad
Yanzhao Zhou\inst{1}\thanks{Corresponding Author.}\\
}

\authorrunning{F.~Author et al.}

\institute{University of Chinese Academy of Sciences \\
\email{\ liufeng20@mails.ucas.ac.cn \quad \{wanfang,qxye,zhouyanzhao\}@ucas.ac.cn}
\and Mach Drive\\
\email{\{tengteng.huang,qianjing.zhang,haotian.yao,chi.zhang\}@mach-drive.com}}
\maketitle

\begin{abstract}
    Multi-view 3D object detection systems often struggle with generating precise predictions due to the challenges in estimating depth from images, increasing redundant and incorrect detections. Our paper presents Ray Denoising, an innovative method that enhances detection accuracy by strategically sampling along camera rays to construct hard negative examples. These examples, visually challenging to differentiate from true positives, compel the model to learn depth-aware features, thereby improving its capacity to distinguish between true and false positives.
    Ray Denoising is designed as a plug-and-play module, compatible with any DETR-style multi-view 3D detectors, and it only minimally increases training computational costs without affecting inference speed.
    Our comprehensive experiments, including detailed ablation studies, consistently demonstrate that Ray Denoising outperforms strong baselines across multiple datasets. It achieves a 1.9\% improvement in mean Average Precision (mAP) over the state-of-the-art StreamPETR method on the NuScenes dataset. It shows significant performance gains on the Argoverse 2 dataset, highlighting its generalization capability. The code will be available at \href{https://github.com/LiewFeng/RayDN}{\color{magenta}https://github.com/LiewFeng/RayDN}.

\end{abstract}

\section{Introduction}
\label{sec:intro}

\begin{figure}[t]
  \centering
    \includegraphics[width=0.65\linewidth]{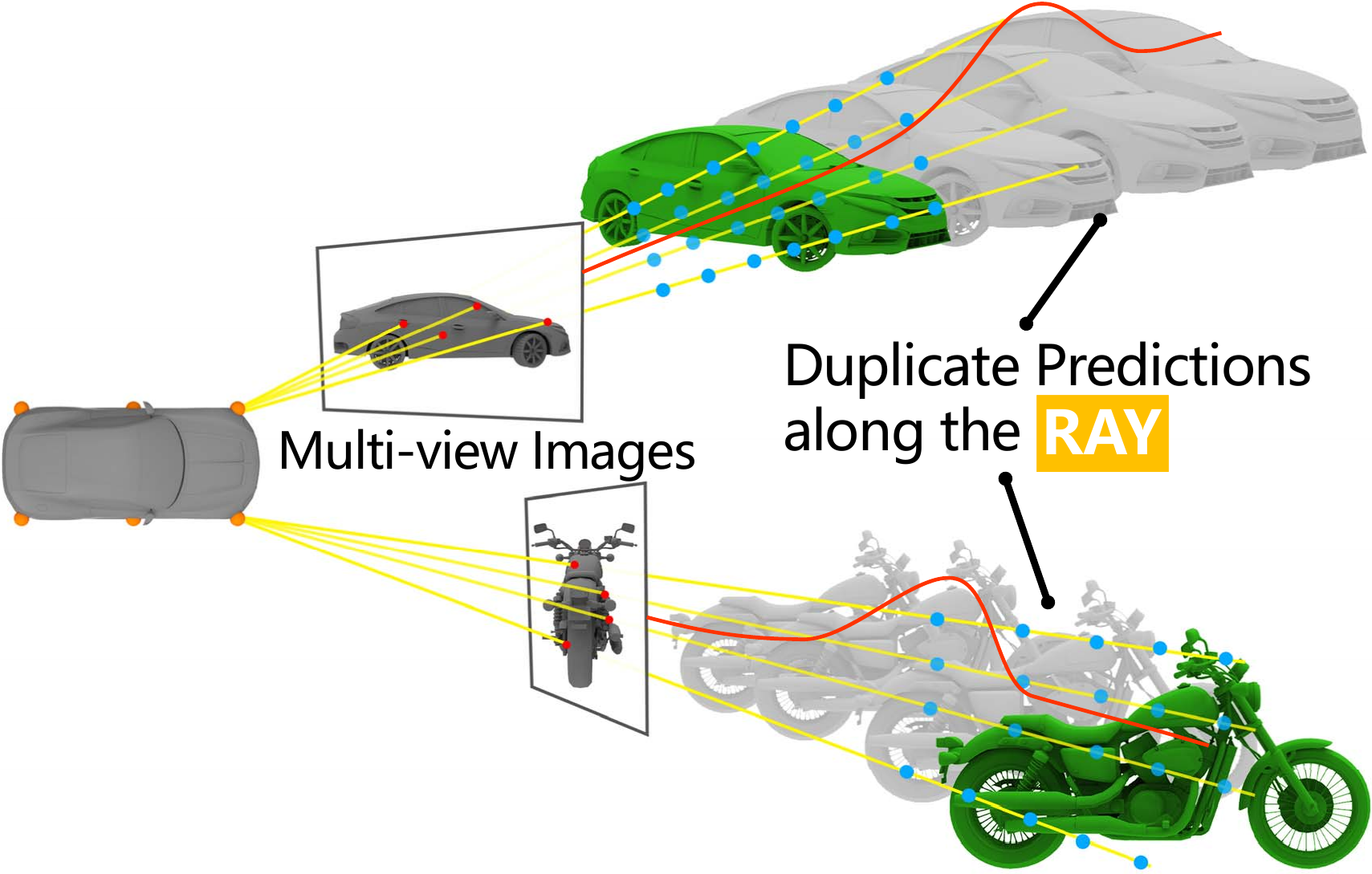}
  \caption{The challenge of estimating depth from images in multi-view 3D object detection leads to duplicate predictions and false positive detections along camera rays. Best viewed in color.}
  \label{fig:teaser}
\end{figure}

3D object detection is a crucial component in autonomous driving systems, drawing considerable interest from the computer vision community. The field of image-based 3D object detection~\cite{huang2021bevdet, li2023bevdepth, li2023bevstereo, wang2022detr3d, jiang2023polarformer, wang2023object} is experiencing a surge in research due to its cost-effectiveness compared to LiDAR-based solutions. A key challenge in multi-view 3D object detection, which relies on images from surrounding cameras, is the difficulty in estimating depth from images, leading to duplicate predictions, as shown in Figure~\ref{fig:teaser}.

Despite the methodological improvements, multi-view 3D object detectors struggle to reduce false positive predictions arising from depth ambiguities. Several recent studies~\cite{li2022bevformer, yang2023bevformer, liu2023petrv2, lin2022sparse4d, Zong_2023_ICCV, liu2023sparsebev, Wang_2023_ICCV, huang2022bevdet4d, park2022time} have attempted to tackle this issue by incorporating temporal information. However, these methods do not explicitly account for the 3D structure of the scene, which limits their potential for further enhancement.

Furthermore, previous works have explored applying general techniques such as Non-Maximum Suppression (NMS) and Focal Loss to mitigate duplicate predictions. NMS, a post-processing technique, targets false positive predictions with high Intersection over Union (IoU) but is less effective when these predictions are scattered along rays with low IoU. Focal Loss, a loss function designed to reduce high-confidence false positive predictions, has also been implemented. However, it has been observed that multi-view 3D object detectors using Focal Loss still face challenges in effectively resolving the problem of false positive predictions along the same rays.

Our quantitative analysis highlighted the importance of addressing false positive predictions along the same ray as Ground Truth. By utilizing the precise positional data of Ground Truth objects, we could identify and eliminate these redundant predictions within the state-of-the-art StreamPETR method~\cite{Wang_2023_ICCV}. This process significantly enhanced $5.4\%$ in mean Average Precision (mAP), emphasizing the critical need for models to improve their depth estimation capabilities. The substantial improvement underscores the potential of refining depth estimation to suppress these predictions and enhance overall detection performance.



\begin{figure}[t]
  \centering
    \includegraphics[width=\linewidth]{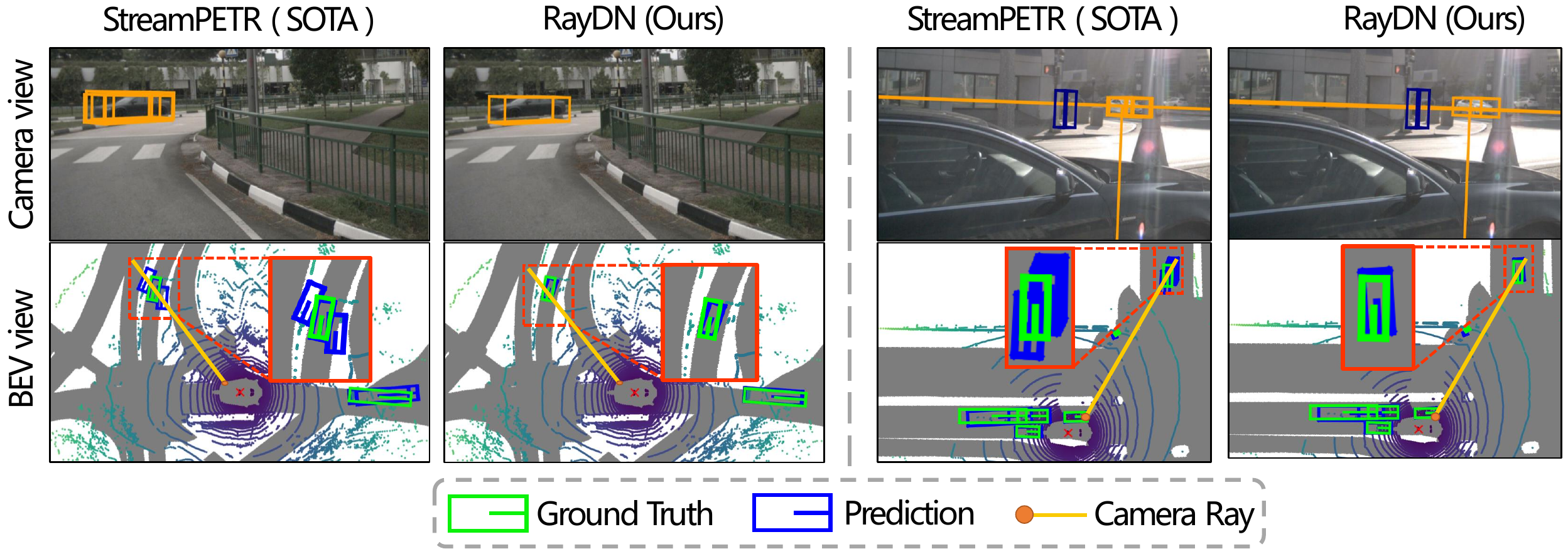}
  \caption{The proposed Ray Denoising approach (right) effectively reduces false positive detections along the ray (highlighted by red rectangles) in the previous state-of-the-art method StreamPETR~\cite{Wang_2023_ICCV} (left). Best viewed by zooming on the screen.}
  \label{fig:motivation}
\end{figure}

Our key observation is that false positives often occur along camera rays due to the inherent limitation in conventional multi-view object detectors. Since the depth information for each pixel is not accurately estimated, the position embedding can only encode the ray direction. As a result, queries on the same ray will consistently interact with identical visual features from the image, leading to numerous duplicate predictions (false positives) along that ray. This scenario underscores the model's need to learn depth-aware features that can discern objects in depth despite the visual features being the same for objects along the same ray.
We propose a novel method called Ray Denoising (\ie, RayDN). This framework is inherently flexible and does not limit the choice of distribution for sampling depth-aware hard negative samples. Based on our ablation studies, we have chosen the Beta distribution for its effectiveness in capturing the spatial distribution of false positives that models are likely to generate. This choice enables Ray Denoising to create depth-aware hard negative samples used for denoising, thereby enhancing the model's ability to learn more robust features and representations for distinguishing false positives along the ray, as depicted in Figure~\ref{fig:motivation}. Ray Denoising introduces only a marginal increase in computational costs during the training stage without affecting inference speed.

To summarize our contributions, we highlight the following key points. Firstly, we have identified the persistent challenge of false positive predictions along the same ray, which acts as a bottleneck in the performance of multi-view 3D object detectors. Secondly, we introduce Ray Denoising, a novel denoising method that utilizes the Beta distribution to create depth-aware hard negative samples along rays. This method explicitly takes into account the 3D structure of the scene, offering a flexible solution compatible with any DETR-style multi-view 3D detector to address the issue of duplicate predictions along rays. Lastly, our method achieves state-of-the-art results on the NuScenes dataset~\cite{caesar2020nuscenes}, significantly enhancing the performance of multi-view 3D object detectors. Specifically, we have improved upon the current state-of-the-art method, StreamPETR, by 1.9\% in mean Average Precision (mAP), thereby demonstrating the effectiveness of Ray Denoising.

\section{Related Work}

\subsection{Image-based 3D Object Detection}
3D object detection is a cornerstone task in autonomous driving systems. The domain has witnessed notable advancements in monocular 3D object detection, largely attributed to the KITTI benchmark~\cite{geiger2013vision}. This progress has been driven by a range of research studies~\cite{simonelli2019disentangling, brazil2019m3d, reading2021categorical, zhang2021objects, wang2021fcos3d, ma2021delving}. Nevertheless, monocular systems face limitations due to their dependence on a single viewpoint and limited data, which hampers their capacity to manage complex scenarios. To overcome these limitations, extensive benchmarks~\cite{caesar2020nuscenes, sun2020scalability, wilson2021argoverse} have been developed, offering multi-viewpoint data that enriches the field of multi-view 3D object detection and propels the evolution of advanced detection methodologies.

The multi-view 3D object detection research is primarily bifurcated into dense Bird's Eye View (BEV)-based and sparse query-based algorithms. Our work is categorized under the sparse query-based algorithms. Dense BEV-based algorithms convert multi-view image features into a dense BEV representation using the Lift-Splat-Shoot (LSS) technique~\cite{philion2020lift}. To mitigate overfitting from LSS, BEVDet~\cite{huang2021bevdet} recommends data augmentation on BEV features. BEVDepth~\cite{li2023bevdepth} introduces depth estimation supervision to enhance LSS precision. Conversely, sparse query-based algorithms such as DETR3D~\cite{wang2022detr3d} and PETR~\cite{liu2022petr} utilize learnable 3D object queries and position encoding to engage with multi-view 2D features. Recent studies~\cite{li2022bevformer, huang2022bevdet4d, liu2023petrv2, lin2022sparse4d, liu2023sparsebev, Wang_2023_ICCV, li2023bevstereo, park2022time} have integrated temporal information to refine these methods further.

\subsection{Hard Negative Samples Mining}
Hard negative sample mining is a strategy to minimize false positive predictions in object detection. Early works like R-CNN~\cite{girshick2015region} employed standard hard negative mining methods~\cite{sung1998example, felzenszwalb2009object}, where misclassified samples were utilized for retraining. OHEM~\cite{shrivastava2016training} and RetinaNet~\cite{lin2017focal} introduced adaptive sampling and focal loss to address the challenges of heuristics and hyper-parameters in hard negative sample selection. These methods select hard negative samples from existing predictions based on confidence scores and matched ground-truth object classes. Our approach takes a different path by constructing new hard negative samples based on the locations of ground-truth objects rather than selecting from existing predictions or depending on prediction confidence and matched ground-truth object classes.

\subsection{Denoising in Object Detection}
In 2D object detection, models such as DETR~\cite{carion2020end} and its variants~\cite{wang2022anchor, liu2022dab, zhu2020deformable} often face convergence issues, which are partly due to the instability of bipartite graph matching and the inconsistency in optimization goals during the early stages of training. To tackle this, DN-DETR~\cite{li2022dn} introduces the concept of feeding noisy ground-truth bounding boxes into the Transformer decoder, with the model being trained to reconstruct the original, clean boxes. This process, known as denoising, involves using queries initialized with these noisy ground-truth bounding boxes, referred to as denoising queries. DINO~\cite{zhang2022dino} further refines this approach by generating both positive and negative denoising queries for each ground-truth bounding box, with the negative queries incorporating more noise to improve the model's ability to reject incorrect predictions.

In the realm of 3D object detection, DETR-style methods~\cite{liu2023petrv2, Wang_2023_ICCV, liu2023sparsebev} perpetuate the denoising approach by creating a denoising query for each ground-truth bounding box. Based on the noise level, these queries are classified as either an object or 'no object'. However, these methods neglect to incorporate knowledge of the 3D scene structure when generating denoising queries. In multi-view 3D object detection, the absence of depth information for each pixel results in models producing duplicate predictions along the same ray with different depths. To combat these false positive predictions, we harness the 3D structure knowledge of the scene. We devise multiple negative ray-denoising queries distributed along the rays, spatially distinct from the traditional negative denoising queries surrounding the ground-truth objects. This innovative design markedly improves the detector's capability to differentiate between true positive predictions and false positive predictions along the same ray.

\section{Methodology}
\label{sec:method}
This section elaborates on our proposed method, Ray Denoising, for multi-view 3D object detection. We start with a comprehensive overview of the DETR-style multi-view 3D object detector framework in Section~\ref{sec: overview}. Following that, we describe the three pivotal steps for integrating Ray Denoising into the framework to address the challenge of duplicate predictions. Section~\ref{sec: generation} outlines the ray generation process. Section~\ref{sec: sampling} explains utilizing the Beta distribution family for sampling reference points. Section~\ref{sec: denoising} details the construction of spatial denoising queries, referred to as Ray Queries, along each ray. Lastly, Section~\ref{sec:discussion} delves into the essential differences between our Ray Queries and the denoising queries employed in prior works and the influence of Ray Denoising on the model training process.

\subsection{Overview}
\label{sec: overview}
Ray Denoising is designed to be integrated into any DETR-style multi-view 3D object detector~\cite{wang2022detr3d, liu2022petr, li2022bevformer, liu2023petrv2, liu2023sparsebev, lin2022sparse4d, Wang_2023_ICCV}. These detectors typically consist of a convolutional network-based feature encoder and a transformer-based decoder. The process begins by feeding $N$ surround-view images $\boldsymbol{I}=\{I_{i} \in \mathbb{R}^{3\times H_{I} \times W_{I}}, i=1, 2, ..., N\}$ into the feature encoder (e.g., ResNet101~\cite{he2016deep}) to extract image features $\boldsymbol{F}=\{F_{i} \in \mathbb{R}^{C\times H_{F} \times W_{F}}, i=1, 2, ..., N\}$. Here, $H_{I}$ and $W_{I}$ denote the image dimensions, $H_{F}$ and $W_{F}$ are the feature dimensions, and $C$ represents the number of channels in the feature. To facilitate 3D perception, multiple points within each camera's frustum are transformed and encoded into position embeddings~\cite{liu2022petr}. These embeddings enable the multi-view image features to interact with 3D queries.

The final step involves $N$ object queries $\mathbf{Q}\in\mathbb{R}^{N \times 256}$, which are derived from a set of learnable 3D reference points $\mathbf{P}\in\mathbb{R}^{N \times 3}$. These queries interact with the multi-view image features $\boldsymbol{F}$ in the transformer decoder, employing multi-layer cross-attention to identify objects. The decoder's output features are then processed by prediction heads (multi-layer perceptron, \ie, MLP) to yield classification scores ($cls$), position offsets ($x, y, z$), scales ($w, h, l$), directions ($\theta_x, \theta_y$), and velocities ($v_x, v_y$).

\begin{figure*}[t]
  \centering
    \includegraphics[width=0.98\linewidth]{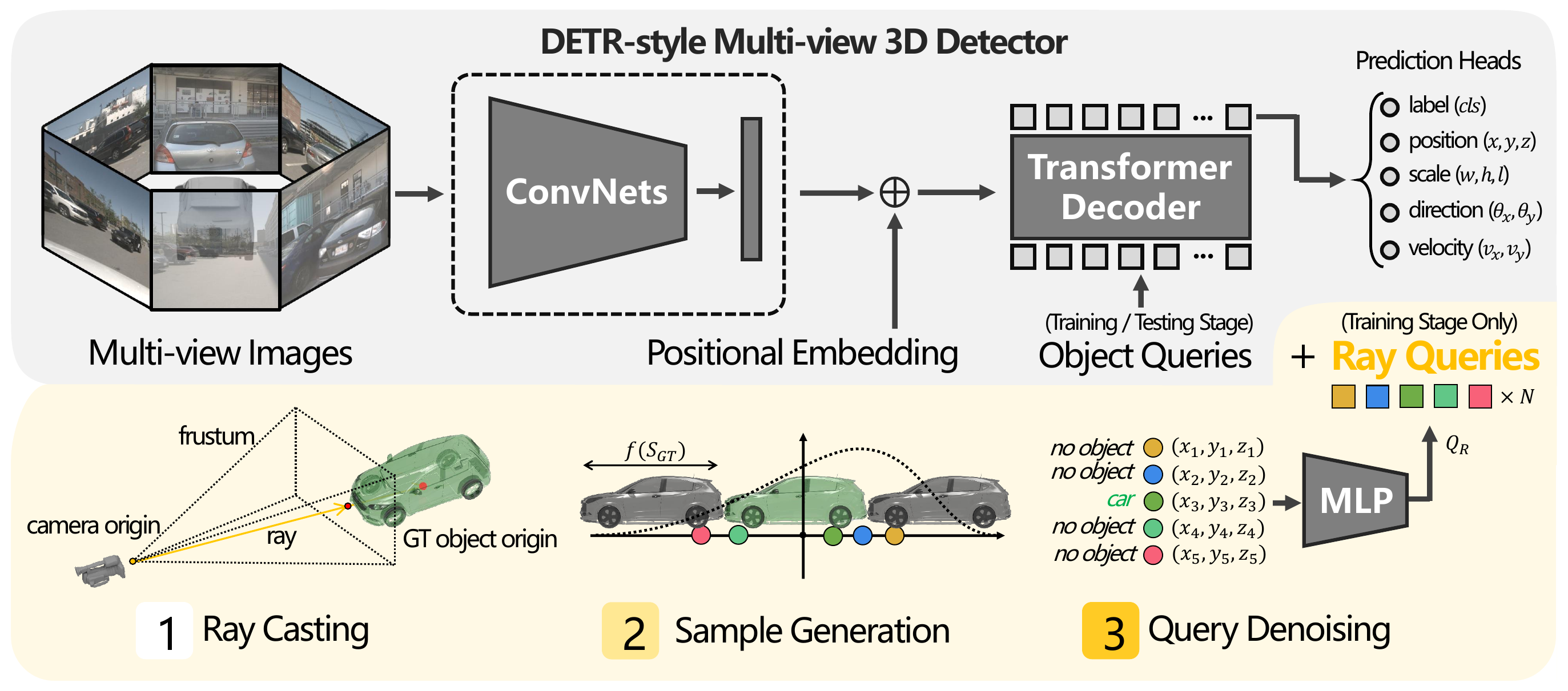}
  \caption{Overall framework of the Ray Denoising approach, a plug-and-play training technique for DETR-style multi-view 3D object detectors, focuses on refining the model's ability to distinguish true positives from false positives in depth. Casting rays and sampling depth-aware denoising queries effectively tackle the challenge of false positives arising from the inherent difficulties in visually estimating depth, leading to substantial improvements in detection performance over strong baselines. Best viewed in color and by zooming on the screen.}
  \label{fig:framework}
\end{figure*}

\subsection{Ray Casting}
\label{sec: generation}
The reference points for our ray denoising queries are distributed along the camera ray, which extends from the camera's optical center to the ground-truth object on the image plane. To establish this ray, we project the 3D center of the ground-truth object into the camera frustum space using the following transformation:
\begin{equation}
    \label{eq:project}
        \mathbf{C'} = \mathbf{K} \cdot \mathbf{C}_{GT},
\end{equation}
where $\mathbf{K}$ is the $4 \times 4$ transformation matrix that maps points from 3D world space to camera frustum space. $\mathbf{C}_{GT} = (x, y, z, 1)$ is the center of the ground-truth object in 3D space, and $\mathbf{C'} = (u \times d, v \times d, d, 1)$ is the corresponding projected center in the camera frustum, with $(u, v)$ being the pixel coordinates and $d$ the depth value.

With the depth of the ground-truth object's center known, we can determine the coordinates of valid reference points along the ray as follows:
\begin{equation}
    \label{eq:camera-ray denoising}
        \hat{d}_{i} = d + \beta_{i} \cdot f(\mathbf{S}_{GT}),
\end{equation}
where $f$ is a function that encodes the average scale of the ground-truth object, calculated as $f(\mathbf{S}_{GT}) = k \cdot \frac{w + h + l}{6}$. The radius $k$ defines the valid distribution range for reference points, and $\beta_{i}$ is the offset for the $i$-th reference point.

The position of the reference point is then obtained by:
\begin{equation}
    \label{eq:project back}
        \hat{\mathbf{P}_{i}} = \mathbf{K}^{-1} \cdot \hat{\mathbf{C'}_{i}},
\end{equation}
where $\hat{\mathbf{C'}_{i}} = (u \times \hat{d}_{i}, v \times \hat{d}_{i}, \hat{d}_{i}, 1)$.

\begin{figure}[t]
	\centering
	\begin{subfigure}{0.49\linewidth}
		\centering
		\includegraphics[width=\linewidth]{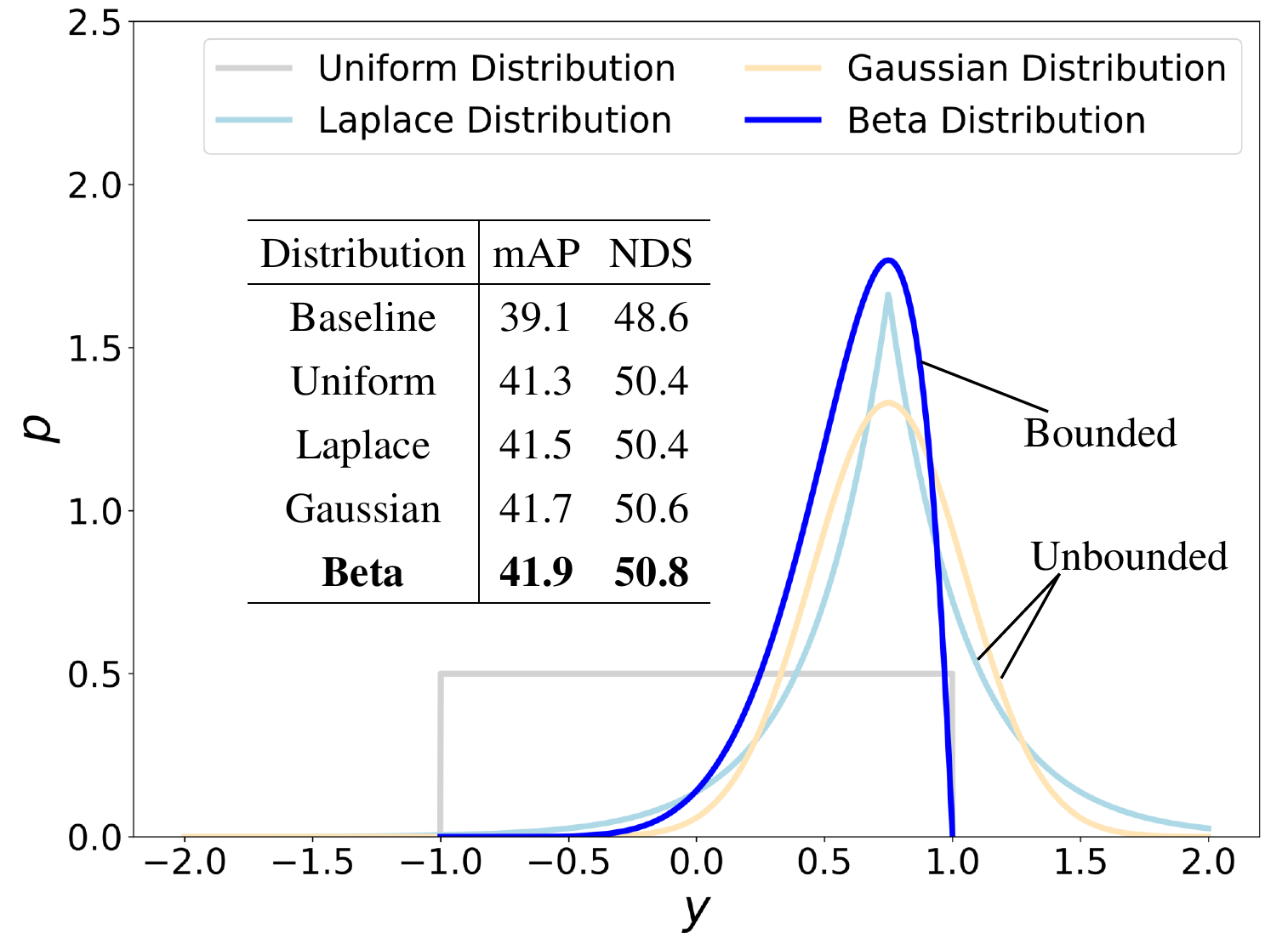}
		\caption{Distribution Comparison}
            \label{fig:distribution}
	\end{subfigure}
	\centering
	\begin{subfigure}{0.49\linewidth}
		\centering
		\includegraphics[width=\linewidth]{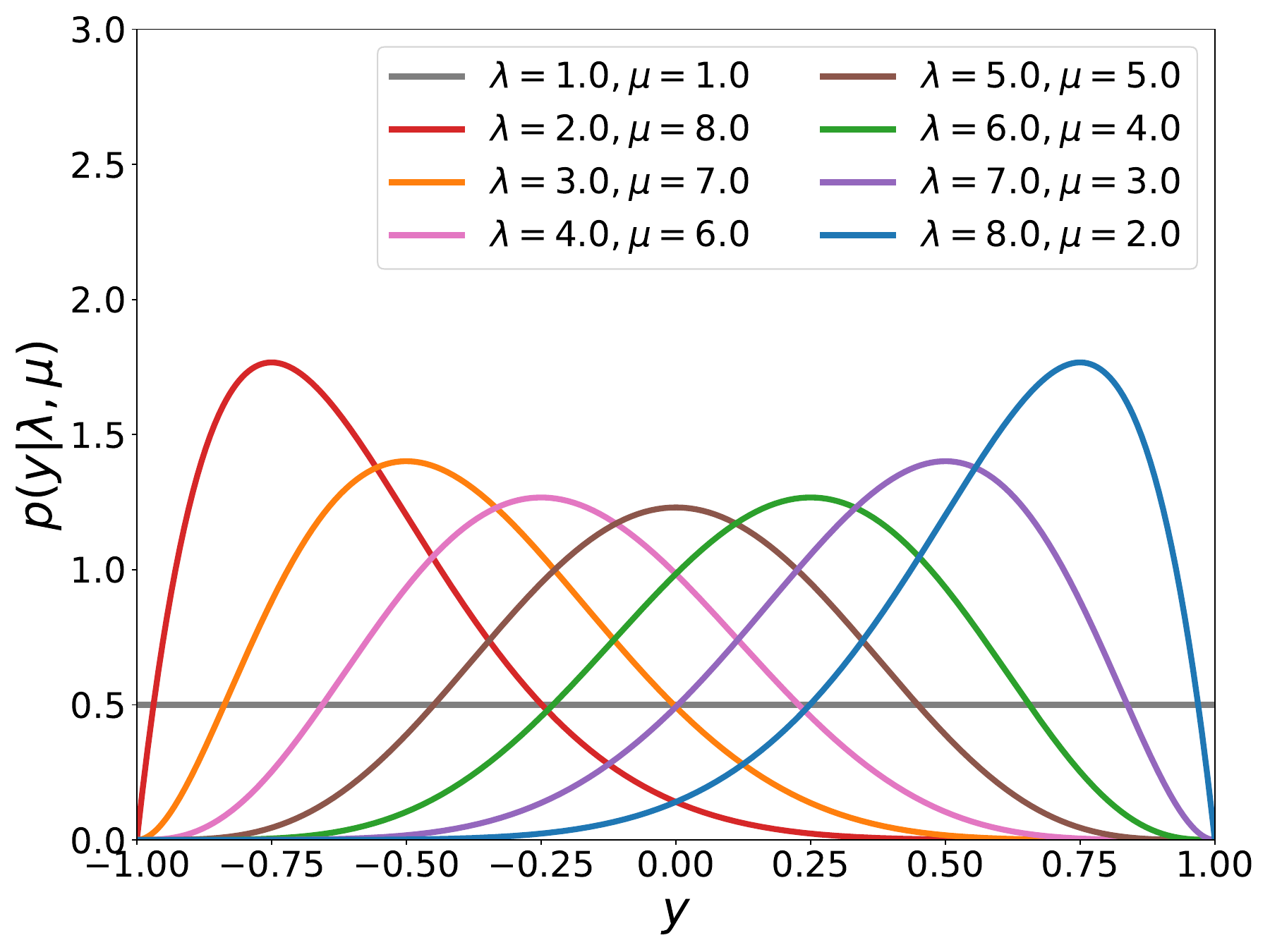}
		\caption{Beta distribution family}
            \label{fig:density}
	\end{subfigure}
	\caption{(a) Distribution comparison showing that the Beta distribution is bounded between -1 and 1, unlike the Laplace and Gaussian distributions, which are unbounded. (b) The Beta distribution family, with the x-range adjusted from $[0,1]$ to $[-1,1]$ using the transformation $y=2x-1$. Best viewed in color.}
\end{figure}

\subsection{Sample Generation}
\label{sec: sampling}
With the valid reference points along the rays defined, we proceed to sample $N$ reference points to emulate the distribution of false positive predictions that arise from depth ambiguities. A straightforward uniform distribution has proven effective, yielding a 2.2\% mAP improvement as depicted in Figure~\ref{fig:distribution}. To more accurately simulate the distribution of false positives, we consider two key aspects. First, we assess whether models tend to predict objects as being closer or further away from the ego-vehicle relative to the actual positions of the ground-truth objects. Second, we examine whether models are more likely to predict objects near or far from the center of the ground-truth objects.

Distributions like the Beta, Gaussian, and Laplace are suitable for handling complex scenarios, with the Gaussian and Laplace being unbounded. However, samples with significant offsets from the ground truth can result in large regression losses, negatively impacting optimization. As shown in Figure~\ref{fig:distribution}, the Beta distribution slightly outperforms the others. Notably, the uniform distribution is a special case within the Beta family. For a versatile sampling strategy applicable to all scenarios, we select the Beta distribution family, characterized by the probability density function:
\begin{equation}
    \label{eq:beta}
        p(x|\lambda,\mu) = \frac{\Gamma(\lambda + \mu)}{\Gamma(\lambda) + \Gamma(\mu)} x^{\lambda - 1}(1 - x)^{\mu - 1},
\end{equation}
where $\lambda$ and $\mu$ are hyper-parameters that shape the distribution. The Gamma function, denoted as $\Gamma(x)$, is defined as:
\begin{equation}
    \label{eq:gamma}
        \Gamma(x) = \int_{0}^{\infty} t^{x - 1} e^{-t} dt.
\end{equation}

The beta distribution's probability density function (PDF), visualized in Figure~\ref{fig:density}, reveals its behavior under different parameter settings. When $\lambda$ equals $\mu$, the distribution is symmetric, reflecting a scenario where the model predicts objects with equal likelihood at closer or further distances. For instance, with $\lambda = \mu = 1$, the distribution simplifies to a uniform distribution. Adjusting $\lambda$ and $\mu$ to be higher shifts the reference points of the denoising queries closer to the ground-truth object's center and vice versa.

When $\lambda$ exceeds $\mu$, such as in the case of $\lambda = 8$ and $\mu = 2$, the distribution suggests that models are more likely to predict objects further away from the ego-vehicle. Conversely, when $\lambda$ is less than $\mu$, for example, $\lambda = 2$ and $\mu = 8$, the distribution indicates a propensity for models to predict objects closer to the ego-vehicle. This flexibility in the Beta distribution allows for tailored sampling strategies that can adapt to the specific challenges of depth estimation in multi-view 3D object detection.

\subsection{Query Denoising}
\label{sec: denoising}
We construct ray denoising queries from the sampled $N$ reference points. Specifically, a multi-layer perceptron (MLP) is utilized to project the normalized 3D coordinates of each point into a latent feature space. These $N$ ray denoising queries $\boldsymbol{q}_r$ are combined with the learnable object queries $\boldsymbol{q}_o$ from the baseline detector and are input into the transformer decoder. An attention mask is employed during the self-attention operations to prevent information leakage from the ray denoising queries to the object queries. The query closest to the center of the ground truth (GT) object is designated as the positive query, inheriting the class label of the corresponding GT object. The remaining $N-1$ queries are labeled as background, \ie, 'no object'. The loss for ray denoising queries adheres to the same criteria as for the learnable object queries, using Focal Loss for classification and L1 Loss for regression. This process ensures that the model is effectively trained to discern between true positives and false positives along the same ray.

\subsection{Discussion}
\label{sec:discussion}
Our Ray Denoising approach is based on the pivotal observation that image-based 3D detection systems often struggle to distinguish true positives from false positives along camera rays. DETR-style multi-view 3D object detectors implicitly learn depth estimation from ground truth supervision. However, the randomly distributed reference points of learnable queries do not fully leverage the available ground truth information. While these reference points are updated during training, they fail to provide adequate hard negative samples for each object in every scene.
To enhance the utilization of ground truth information, traditional denoising techniques introduce additional reference points distributed uniformly around the ground-truth object during training. These instance-specific reference points have improved detection performance~\cite{Wang_2023_ICCV, liu2023petrv2, liu2023sparsebev}. However, they overlook the depth ambiguities intrinsic to multi-view 3D object detection. The absence of precise depth information for each pixel means that the position embedding can only encode ray direction, not depth. This leads to queries on the same ray interacting with the same image features, resulting in redundant predictions.
Ray Denoising diverges from traditional techniques by strategically generating reference points along rays that extend from cameras to objects. This approach explicitly considers the 3D structure of each object in the scene, providing a sufficient number of hard negative samples. During training, these Ray Queries interact within the self-attention layer, effectively guiding the model to suppress depth-ambiguous spatial hard negative samples. This interaction enhances the detector's ability to differentiate between true positive predictions (objects) and false positive predictions (duplicate detections), improving detection accuracy.

\section{Experiment}
\label{sec:exp}

\subsection{Dataset and Metrics}
Our model's performance was assessed using two datasets: nuScenes~\cite{caesar2020nuscenes} and Argoverse 2~\cite{wilson2021argoverse}. The nuScenes dataset include 1000 video sequences. which are split into training (700 videos), validation (150 videos), and testing (150 videos) sets, each approximately 20 seconds long with annotations at 0.5-second intervals. The dataset comprises 1.4 million annotated 3D bounding boxes across ten object classes. Evaluation metrics include the mean Average Precision (mAP) and five true positive metrics: ATE, ASE, AOE, AVE, and AAE, which assess translation, scale, orientation, velocity, and attribute errors. The nuScenes Detection Score (NDS) is a comprehensive score derived from these metrics, offering an overall performance evaluation.

The Argoverse 2 dataset contains 1000 unique scenes, each 15 seconds long, annotated at 10 Hz. The scenes are divided into 700 for training, 150 for validation, and 150 for testing. The evaluation covers 26 categories within a 150-meter range, addressing long-range perception tasks. Metrics include the mAP and the Composite Detection Score (CDS), which integrates three other true positive metrics: ATE, ASE, and AOE.

\subsection{Implementation Details}
Our experimental setup leverages ResNet50~\cite{he2016deep}, ResNet101, and V2-99~\cite{lee2019energy} backbones, each with distinct pre-training configurations. We detail the performance of ResNet50 and ResNet101 models, which are pre-trained on nuImages~\cite{caesar2020nuscenes}, on the nuScenes validation set. To showcase the scalability of our approach, we also report results on the nuScenes test set using V2-99, initialized from the DD3D~\cite{park2021pseudo} checkpoint.
Our models are optimized using the AdamW~\cite{loshchilov2018decoupled} optimizer with a batch size of 16. The learning rate is set to $4 \times 10^{-4}$ for models trained only on the training set and $3 \times 10^{-4}$ for those trained on both the training and validation sets. We adopt a cosine annealing policy for learning rate scheduling. For benchmarking against state-of-the-art (SOTA) methods, models are trained for 60 epochs without CBGS~\cite{zhu2019class} and for 24 epochs in ablation studies in nuScenes. For experiments on Argoverse 2, models are trained for 6 epochs. Our implementation is primarily based on the StreamPETR~\cite{Wang_2023_ICCV} framework.
%

\begin{table*}[t]
  \scriptsize          
  \centering
\begin{tabular}{l | c | c | cc | cccccc}
\toprule
\textbf{Methods}     & \textbf{Backbone}      & \textbf{Image Size}  & \textbf{mAP}  &\textbf{NDS}      & \textbf{mATE}    & \textbf{mASE}          & \textbf{mAOE}       & \textbf{mAVE}     & \textbf{mAAE} \\
\midrule
BevDet4D~\cite{huang2022bevdet4d}    & ResNet50      & 256$\times$704                             & 32.2                   & 45.7                   & 0.703                    & 0.278                    & 0.495                    & 0.354                    & 0.206                    \\
PETRv2~\cite{liu2023petrv2}      & ResNet50      & 256$\times$704                            & 34.9                   & 45.6                   & 0.700                      & 0.275                    & 0.580                     & 0.437                    & 0.187                    \\
BEVDepth~\cite{li2023bevdepth}    & ResNet50      & 256$\times$704                            & 35.1                   & 47.5                   & 0.629                    & 0.267                    & 0.479                    & 0.428                    & 0.198                    \\
BEVStereo~\cite{li2023bevstereo}   & ResNet50      & 256$\times$704                        & 37.2                   & 50.0                     & 0.598                    & 0.270                     & 0.438                    & 0.367                    & 0.190                     \\
BEVFormerv2\cite{yang2023bevformer}$\dagger$ & ResNet50      & -                & 42.3                   & 52.9                   & 0.618                    & 0.273                    & 0.413                    & 0.333                    & 0.188                    \\
SOLOFusion~\cite{park2022time}  & ResNet50      & 256$\times$704       & 42.7                   & 53.4                   & 0.567                   & 0.274                    & 0.511                    & 0.252                    & 0.181                   \\
SparseBEV~\cite{liu2023sparsebev}$\dagger$   & ResNet50      & 256$\times$704                             & 44.8                   & 55.8                   & 0.581                    & 0.271                    & 0.373                  & 0.247                   & 0.190                     \\
StreamPETR~\cite{Wang_2023_ICCV}$\dagger$  & ResNet50      & 256$\times$704                           & 45.0                    & 55.0                    & 0.613                    & 0.267                    & 0.413                    & 0.265                    & 0.198                    \\
\rowcolor{gray0}  RayDN$\dagger$ (Ours)        & ResNet50      & 256$\times$704   &\textbf{46.9} & \textbf{56.3} & 0.579 & 0.264 & 0.433 & 0.256 & 0.187 \\
\midrule
BEVDepth~\cite{li2023bevdepth}    & ResNet101     & 512$\times$1408                           & 41.2                   & 53.5                   & 0.565                    & 0.266                    & 0.358                    & 0.331                    & 0.190                     \\
PETRv2~\cite{liu2023petrv2}$\dagger$        & ResNet101 & 640$\times$1600                            & 42.1                   & 52.4                   & 0.681                    & 0.267                    & 0.357                    & 0.377                    & 0.186                     \\
Sparse4D~\cite{lin2022sparse4d}$\dagger$    & ResNet101 & 900$\times$1600                             & 43.6                   & 54.1                   & 0.633                    & 0.279                    & 0.363                    & 0.317                    & 0.177                    \\
SOLOFusion~\cite{park2022time}  & ResNet101     & 512$\times$1408     & 48.3                   & 58.2                   & 0.503                    & 0.264                    & 0.381                    & 0.246                    & 0.207                    \\
SparseBEV~\cite{liu2023sparsebev}$\dagger$   & ResNet101     & 512$\times$1408                           & 50.1                   & 59.2                   & 0.562                    & 0.265                    & 0.321                    & 0.243                    & 0.195                    \\
StreamPETR~\cite{Wang_2023_ICCV}$\dagger$  & ResNet101     & 512$\times$1408                            & 50.4                   & 59.2                   & 0.569                    & 0.262                    & 0.315                    & 0.257                    & 0.199                    \\
\rowcolor{gray0} RayDN$\dagger$ (Ours)        & ResNet101     & 512$\times$1408                          & \textbf{51.8} & \textbf{60.4} & 0.541 & 0.260 & 0.315 & 0.236 & 0.200  \\
\bottomrule
\end{tabular}
\caption{Comparison on the nuScenes validation set. $\dagger$ Indicates methods that benefit from perspective-view pre-training.}
\label{tab:val}
\end{table*}

\begin{table*}[t]
  \scriptsize
  \centering
\begin{tabular}{l | c | c | cc | ccccccc}
\toprule
\textbf{Methods}      & \textbf{Backbone}   & \textbf{Image Size}   & \textbf{mAP}  &\textbf{NDS}      & \textbf{mATE}    & \textbf{mASE}          & \textbf{mAOE}       & \textbf{mAVE}     & \textbf{mAAE}     \\
\midrule
DETR3D~\cite{wang2022detr3d}              & V2-99      & 900$\times$1600         & 41.2                     & 47.9                     & 0.641                     & 0.255                     & 0.394                     & 0.845                     & 0.133                     \\
MV2D~\cite{wang2023object}                & V2-99      & 640$\times$1600         & 46.3                     & 51.4                     & 0.542                     & 0.247                     & 0.403                     & 0.857                     & 0.127                     \\
BEVFormer~\cite{li2022bevformer}           & V2-99      & 900$\times$1600         & 48.1                     & 56.9                     & 0.582                     & 0.256                     & 0.375                     & 0.378                     & 0.126                     \\
PETRv2\cite{liu2023petrv2}              & V2-99      & 640$\times$1600         & 49.0                      & 58.2                     & 0.561                     & 0.243                     & 0.361                     & 0.343                     & 0.120                      \\
PolarFormer~\cite{jiang2023polarformer}         & V2-99      & 900$\times$1600         & 49.3                     & 57.2                     & 0.556                     & 0.256                     & 0.364                     & 0.439                     & 0.127                     \\
BEVStereo~\cite{li2023bevstereo}           & V2-99      & 900$\times$1600         & 52.5                     & 61.0                      & 0.431                    & 0.246                     & 0.358                     & 0.357                     & 0.138                     \\
HoP~\cite{Zong_2023_ICCV}           & V2-99      & 640$\times$1600         & 52.8                     & 61.2                      & 0.491                     & 0.242                     & 0.332                     & 0.343                     & 0.109                     \\
SparseBEV~\cite{liu2023sparsebev}           & V2-99      & 640$\times$1600         & 54.3                     & 62.7                     & 0.502                     & 0.244                     & 0.324                     & 0.251                     & 0.126                     \\
StreamPETR~\cite{Wang_2023_ICCV}          & V2-99      & 640$\times$1600         & 55.0                      & 63.6                     & 0.479                     & 0.239                     & 0.317                     & 0.241                     & 0.119                     \\
\rowcolor{gray0}  RayDN (Ours)                & V2-99      & 640$\times$1600         &\textbf{56.5}      &\textbf{64.5}      & 0.461     & 0.241     & 0.322     & 0.239     & 0.114     \\
\bottomrule
\end{tabular}
\caption{Comparison on the nuScenes test set. 
}
\label{tab:test}
\end{table*}

\subsection{Comparison with State-of-the-Art Methods}
We compare the proposed Ray Denoising method with other state-of-the-art multi-view 3D object detectors on the validation and test sets of the nuScenes dataset, as well as the validation set of the Argoverse 2 dataset. It's important to note that our method does not employ test time augmentation (TTA).

\noindent \textbf{nuScenes Validation Set.} Table~\ref{tab:val} presents a comparison with state-of-the-art methods on the nuScenes validation set. We evaluate both ResNet-50 and ResNet-101 backbones. With ResNet-50 as the backbone and an image size of $704 \times 256$, we achieve $46.9\%$ mAP and $56.3\%$ NDS, improving upon the previous state-of-the-art method, StreamPETR, by $1.9\%$ mAP and $1.3\%$ NDS. Using a more robust ResNet-101 backbone and increasing the image size to $512 \times 1408$, our performance reaches $51.6\%$ mAP and $59.8\%$ NDS, outperforming StreamPETR by $1.2\%$ mAP and $0.6\%$ NDS. These results demonstrate the scalability of Ray Denoising.

\noindent \textbf{nuScenes Test Set.} The results evaluated by the test server are detailed in Table~\ref{tab:test}. Our method includes training on both the training and validation sets. Notably, we achieve $56.5\%$ mAP and $64.5\%$ NDS, surpassing StreamPETR by an absolute $1.5\%$ mAP and $0.9\%$ NDS.

\noindent \textbf{Argoverse 2 Validation Set.} To assess the generalization capability of Ray Denoising, we conduct additional experiments on the Argoverse 2 dataset, as displayed in Table~\ref{tab:argoverse}. Our method significantly outperforms previous state-of-the-art methods, achieving a $2.0\%$ mAP and $1.5\%$ CDS improvement on the validation set. These metrics underscore the generalization capability of our approach.


\begin{table*}[t]
  
  \centering
\begin{tabular}{l  | c | c | cc | ccccc}
\toprule
\textbf{Methods}    & \textbf{Backbone}   & \textbf{Image Size}     & \textbf{mAP}  &\textbf{CDS}      & \textbf{mATE}    & \textbf{mASE}          & \textbf{mAOE}            \\
\midrule
PETR~\cite{liu2022petr} & V2-99 & 900$\times$640  & 17.6  & 12.2 & 0.911  & 0.339  & 0.819 \\
Sparse4Dv2~\cite{lin2023sparse4d} & V2-99  & 900$\times$640 & 18.9 & 13.4 & 0.832 & 0.343 & 0.723 \\
StreamPETR~\cite{Wang_2023_ICCV}  & V2-99  & 900$\times$640 & 20.3 & 14.6 & 0.843 & 0.321 & 0.650 \\
\rowcolor{gray0}  RayDN (Ours) & V2-99      & 900$\times$640 &\textbf{22.3}      &\textbf{16.1}      & 0.825     & 0.325     & 0.629         \\

\bottomrule
\end{tabular}
\caption{Comparisons on the Argoverse 2 validation set. We evaluate across 26 object categories within a range of 150 meters.}
\label{tab:argoverse}
\end{table*}

The tables demonstrate that our Ray Denoising approach notably enhances mAP. Considering that mAP is significantly impacted by incorrect false positives, this enhancement firmly validates the effectiveness of Ray Denoising in reducing redundant predictions along the camera rays.

\subsection{Ablation Study}
This section delves into the ablation studies performed using the validation sets from the nuScenes and Argoverse 2 datasets. Unless specified otherwise, our experiments on nuScenes are based on a ResNet50 backbone, which is pre-trained on the nuImages dataset~\cite{caesar2020nuscenes}. Our input data comprises 8 frames, each with a $704 \times 256$ pixels resolution. The decoder utilizes 428 queries, and the model is trained for 24 epochs, forgoing Class-Balanced Group Sampling (CBGS)~\cite{zhu2019class}.

\noindent \textbf{Radius of Ray Denoising Queries.}
Table~\ref{tab:radius of dn query} shows how varying the radius $k$ affects performance. We find that a radius of $k=3$ yields the optimal results, while $k=2$ and $k=4$ also lead to notable enhancements in performance.

\noindent \textbf{Number of Ray Denoising Queries.}
We explore the effect of the quantity of ray denoising queries on the model's performance in Table~\ref{tab:number of ray dn query}. The findings reveal that both mAP and NDS metrics improve with increasing ray queries, reaching a plateau of 5 queries. Surprisingly, upping the count to 7 queries results in a dip in mAP, which might be attributed to an imbalance in the ratio of positive to negative ray denoising queries.

\noindent \textbf{Distribution of Ray Denoising Queries.}
We investigate the impact of the distribution of ray denoising queries with various hyper-parameters across the nuScenes and Argoverse 2 datasets. The results are detailed in Table~\ref{tab:distribution of ray dn query} and Table~\ref{tab:distribution of ray dn query on argoverse}. Our findings indicate that employing a uniform distribution, which is dataset-agnostic (\ie, setting $\lambda=1$ and $\mu=1$), leads to substantial performance gains for Ray Denoising. Specifically, we observe a 2.2\% increase in mAP for the nuScenes dataset and a 1.0\% increase in mAP for the Argoverse 2 dataset, highlighting the method's strong generalization capabilities. Fine-tuning these hyper-parameters further enhances the performance, with an additional 0.6\% mAP improvement on nuScenes and 1.0\% mAP on Argoverse. Across both datasets, models configured with $\lambda>\mu$ slightly outperform those with $\lambda<\mu$, suggesting that false positives tend to be at greater distance from the ego-vehicle.
\begin{table*}[t]
  \scriptsize
  \begin{floatrow}
  \capbtabbox{
  \centering
  \begin{tabular}{cccccccccccc}
    \toprule
    \textbf{Method} &\textbf{Radius}    &\textbf{mAP}  &\textbf{NDS}                       \\
    \midrule
    SOTA Baseline~\cite{Wang_2023_ICCV} &-   & 39.1                & 48.6                                              \\\hline
    \multirow{3}{*}{+RayDN (Ours)} &2        & 41.1 & 49.8                \\
    &3     & \textbf{41.9} & \textbf{50.8}                   \\
    &4     & 41.3 & 50.1                   \\
    \bottomrule
  \end{tabular}
  }
  {
  \caption{Ablation studies on the radius of ray denoising queries.}
  \label{tab:radius of dn query}
  }
  \capbtabbox{
  \begin{tabular}{cccccccccccc}
    \toprule
     \textbf{Method} &\textbf{\#Q}     &\textbf{mAP}  &\textbf{NDS}                      \\
    \midrule
    SOTA Baseline~\cite{Wang_2023_ICCV} & -  & 39.1                & 48.6                                              \\ \hline
    \multirow{3}{*}{+RayDN (Ours)} &3     & 41.6 & 50.2                  \\
    &5     & \textbf{41.9} & \textbf{50.8}                 \\
    
    &7     &   41.1  & 50.5              \\

    \bottomrule
  \end{tabular}
  }
  {
  \caption{Ablation studies on the number of ray denoising queries.}
  \label{tab:number of ray dn query}
  }
  \end{floatrow}
\end{table*}


    


\begin{table*}[t]
  \scriptsize
  \begin{floatrow}
  \capbtabbox{
  \centering
  \begin{tabular}{cccccccccccc}
    \toprule
    \textbf{Method} &\textbf{$\lambda$} & \textbf{$\mu$}     &\textbf{mAP}  &\textbf{NDS}                 \\
    \midrule
    SOTA Baseline~\cite{Wang_2023_ICCV} &- & -  & 39.1                & 48.6                                      \\\hline
    \multirow{6}{*}{+RayDN (Ours)}&1 & 1     & 41.3 & 50.4                     \\
    &2 & 8     & 41.5 & 50.1                  \\
    &3 & 7     & 41.3 & 50.5 \\
    &7 &3      &41.7 & 50.7  \\
    &8 &2     & \textbf{41.9} & \textbf{50.8}                  \\
    &9 &1     & 41.6  & 50.4 \\
    
    \bottomrule
  \end{tabular}
  }
  {
  \caption{Ablation studies on the distribution of ray denoising queries on nuScenes dataset.}
  \label{tab:distribution of ray dn query}
  }
  \capbtabbox{
  \begin{tabular}{cccccccccccc}
    \toprule
    \textbf{Method} &\textbf{$\lambda$} & \textbf{$\mu$}     &\textbf{mAP}  &\textbf{CDS}                 \\
    \midrule
    SOTA Baseline~\cite{Wang_2023_ICCV} &- & -  & 20.3                & 14.6                                      \\\hline
    \multirow{6}{*}{+RayDN (Ours)}&1 & 1     & 21.3 & 15.4                     \\
    &3 &7     & 21.7 & 15.8                  \\
    &4 &6     & 21.3 & 15.3                   \\
    &6 &4     & 21.5 & 15.7 \\
    &7 &3     & \textbf{22.3} & \textbf{16.1}                  \\
    &8 &2     & 21.5 & 15.6 \\
    
    \bottomrule
  \end{tabular}
  }
  {
  \caption{Ablation studies on the distribution of ray denoising queries for the Argoverse 2 dataset.}
  \label{tab:distribution of ray dn query on argoverse}
  }
  \end{floatrow}
\end{table*}

\begin{figure}[t]
	\centering
	\begin{subfigure}{0.49\linewidth}
		\centering
		\includegraphics[width=\linewidth]{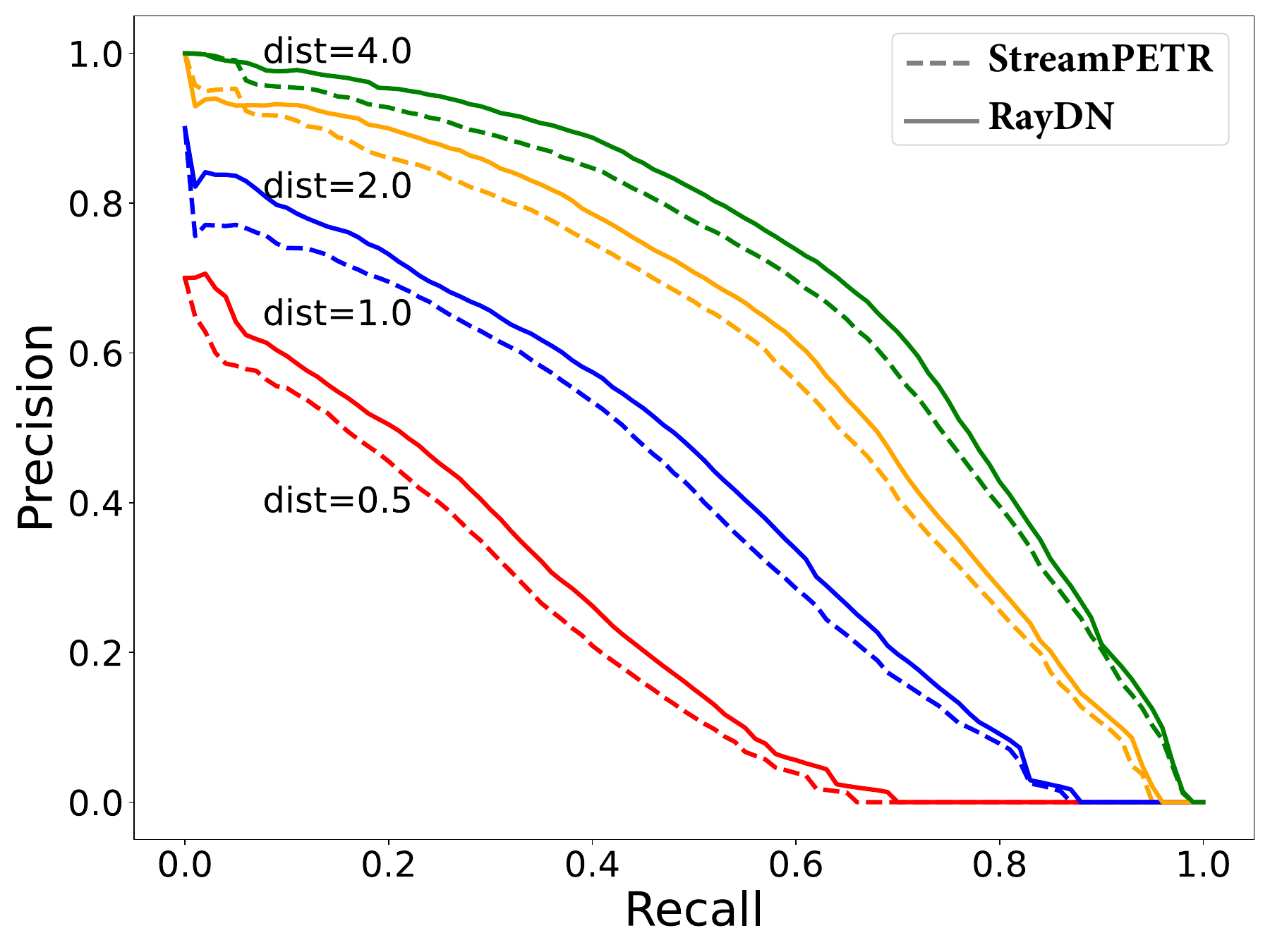}
		\caption{Precision-Recall Curve}
            \label{fig:pr curve}
	\end{subfigure}
	\centering
	\begin{subfigure}{0.49\linewidth}
		\centering
		\includegraphics[width=\linewidth]{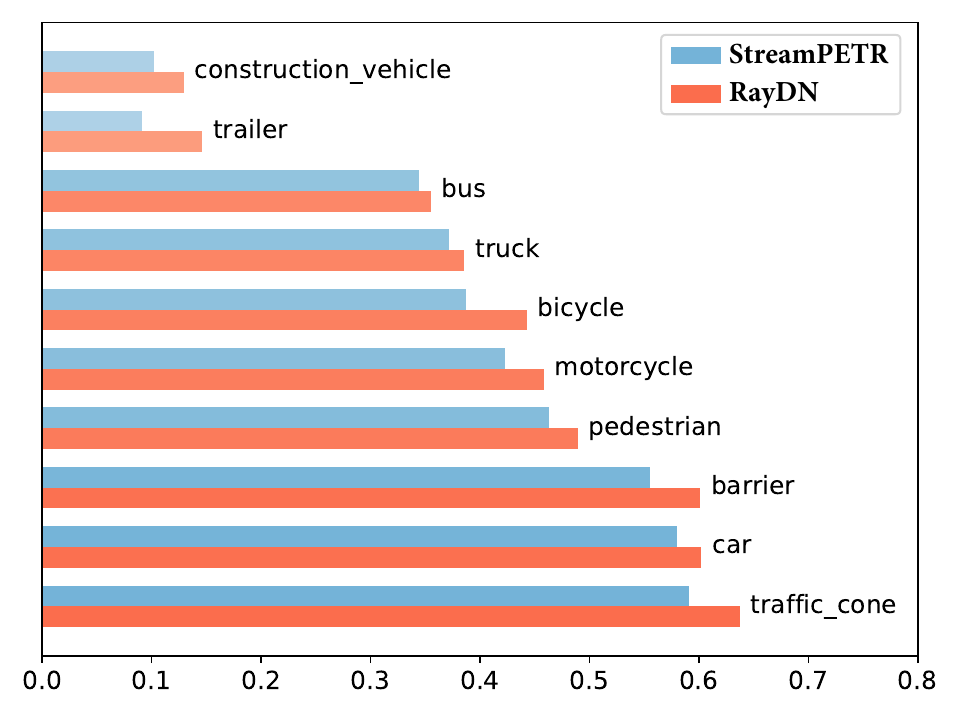}
		\caption{AP for each class}
            \label{fig:ap}
	\end{subfigure}
	\caption{(a) Visualization of the precision-recall curves at various distance thresholds. Ray Denoising consistently enhances precision across nearly all recall levels, effectively suppressing false positives. (b) Class-wise AP comparison. Ray Denoising performs superior over the SOTA StreamPETR in all object classes. Best viewed in color.}
    \label{AP}
\end{figure}

\noindent \textbf{Precision-Recall Analysis.}
We assess the ability of Ray Denoising to reduce false positive predictions by plotting precision-recall curves and Average Precision (AP) for each object class. Figure~\ref{fig:pr curve} reveals that Ray Denoising enhances precision across a range of recall levels and under various distance thresholds. Similarly, Figure~\ref{fig:ap} shows that Ray Denoising surpasses StreamPETR in AP across all classes. These results collectively affirm that Ray Denoising effectively mitigates the occurrence of false positive predictions.

\begin{figure*}[t]
\centering
\includegraphics[width=1.0\linewidth]{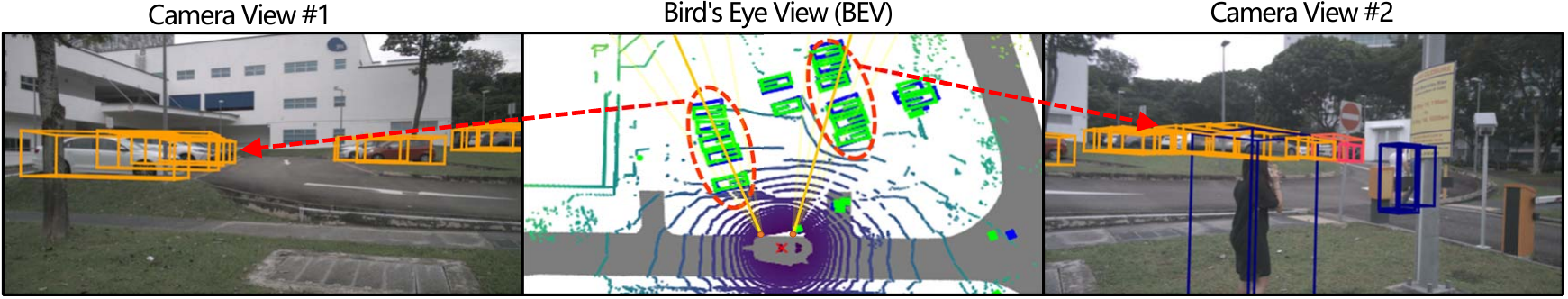}
\caption{Visualization of our detection results on the validation set of nuScenes. Ray Denoising effectively mitigates duplicate false positives while maintaining the ability to detect highly occluded objects along the same ray. Best viewed on the screen.}
\label{fig:detection_results}
\end{figure*}

\noindent \textbf{Visualization of Detection Results.}
Our approach creates multiple spatial hard negative samples along the same ray, potentially interfering with detecting true positives if they are also located along that ray. To test this, we present a visualization of the detection outcomes in Figure~\ref{fig:detection_results}. The results show that Ray Denoising effectively eliminates duplicate false positives without compromising the detection of true positives, even when they are heavily occluded along the same ray.

\noindent \textbf{Generalization Ability of Ray Denoising.}
To assess the versatility of Ray Denoising, we apply it to different models with varied input data. We use the state-of-the-art CMT~\cite{yan2023cross} as a reference and test Ray Denoising with single-frame images and multi-modal data. In the single-frame scenario, we employ a ResNet50 backbone pre-trained on ImageNet~\cite{caesar2020nuscenes}, with input images sized at 256 $\times$ 704 pixels. For the multi-modal setup, the input image resolution is 320 $\times$ 800 pixels, and the voxel dimensions are 0.1m $\times$ 0.1m. The batch size is set to 16 due to memory constraints, and all models are trained for 20 epochs using CBGS~\cite{zhu2019class}. Table~\ref{tab:multi model} shows that Ray Denoising improves upon the baseline with a $1.5\%$ increase in mAP and a $1.4\%$ increase in NDS for the single-frame setup. Moreover, it achieves a $1.0\%$ mAP and a $0.9\%$ NDS enhancement on the already high-performing multi-modal baseline. These experiments confirm the broad applicability of Ray Denoising.

\noindent \textbf{Effect of Depth Estimation on Ray Denoising.}
In DETR-style multi-view 3D object detectors, the lack of depth information leads to the use of camera rays as position embeddings for image features, which can be imprecise. To refine these embeddings, 3DPPE~\cite{Shu_2023_ICCV} suggests encoding an estimated 3D point guided by external supervision for depth estimation.
To explore the role of depth estimation in conjunction with Ray Denoising, we integrate 3DPPE into our framework. The results, detailed in Table~\ref{tab:3dppe}, show a marked improvement in performance with 3DPPE. However, Ray Denoising alone, without external depth supervision, also achieves similar performance. This underscores our core insight: Ray Denoising enhances the model's depth perception by guiding it to learn superior depth perception ability during training. When Ray Denoising is paired with 3DPPE, we further obtain a $+1.9\%$ increase in mAP and a $+1.7\%$ increase in NDS. This demonstrates that while depth estimation algorithms can improve the clarity of position embeddings, Ray Denoising's ability to refine the model's understanding of depth through robust training is a powerful complement, even when depth information is available.

\begin{table*}[t]
  \small
  \begin{floatrow}
  \capbtabbox{
  \centering
  \begin{tabular}{lccccccccccc}
    \toprule
    \textbf{Method}    &\textbf{Description} &\textbf{mAP}  &\textbf{NDS}                      \\
    \midrule

    CMT~\cite{yan2023cross} &single frame     & 30.6  & 37.7                        \\
    + RayDN      &single frame   & \textbf{32.1}     & \textbf{39.1}                                     \\ \hline
    CMT~\cite{yan2023cross} & multi-modal     & 67.6  & 70.4                           \\
    + RayDN      & multi-modal       & \textbf{68.6}     & \textbf{71.3}                  \\
    \bottomrule
  \end{tabular}
  }
  {
  \caption{Ablation studies on the generalization ability of Ray Denoising.}
  \label{tab:multi model}
  }
  \capbtabbox{
  \begin{tabular}{ccccccccc}
    \toprule
    \textbf{3DPPE} &\textbf{RayDN}   &\textbf{mAP}  &\textbf{NDS}                    \\
    \midrule
     &    & 39.1                & 48.6                                             \\
    \cmark &        & 42.0                     & 51.1                                         \\
     & \cmark       & 41.9                     & 50.8                                         \\
     \cmark & \cmark     & \textbf{43.9} & \textbf{52.9}       \\

    \bottomrule
  \end{tabular}
  }
  {
  \caption{Ablation studies on the effect of depth estimation.}
  \label{tab:3dppe}
  }
  \end{floatrow}
\end{table*}

\begin{table}[t]
  \centering
  \begin{tabular}{lccccccccccc}
    \toprule
    \textbf{Method}    &\textbf{backbone} &\textbf{Image Size} &\textbf{Training Time}       & \textbf{FPS}              \\
    \midrule
    SOTA Baseline~\cite{Wang_2023_ICCV}  &ResNet50 &704$\times$256 &  7 h &  10.4             \\
    + 3DPPE~\cite{Shu_2023_ICCV}     &ResNet50 &704$\times$256 & 8.5 h   & 9.9                 \\
    + RayDN (Ours)     &ResNet50 &704$\times$256 & 7.5h    & 10.4               \\
    \bottomrule
  \end{tabular}
  \caption{Ablation studies on the training time and inference speed. 
  }
  \label{tab:cost of dn query}
\end{table}

\noindent \textbf{Cost of Ray Denoising.}
We analyze the computational overhead of Ray Denoising by comparing training times and inference speeds, as detailed in Table~\ref{tab:cost of dn query}. Training time is benchmarked across 8 GeForce RTX 2080 Ti GPUs, while inference speed is measured on a single GeForce RTX 2080 Ti GPU. Our setup utilizes a ResNet50 backbone with an input resolution of $256 \times 704$. Ray Denoising introduces a modest increase in training time—just a $7\%$ rise compared to StreamPETR—while 3DPPE raises it by $21\%$. Inference speed remains on par with StreamPETR, as Ray Denoising is only used in the training phase.

\section{Conclusion}
We introduce Ray Denoising, a method designed to overcome the critical challenge of depth estimation inaccuracy in multi-view 3D object detection. Ray Denoising tackles the issue of false detections along camera rays, which are a direct consequence of imprecise depth information from images.
By leveraging the 3D structure of the scene, Ray Denoising prompts the model to learn depth-aware features, leading to improved differentiation between false and true positives along the same ray without introducing extra inference costs. Our comprehensive experiments on the NuScenes and Argoverse 2 datasets demonstrate that Ray Denoising consistently and significantly outperforms strong baselines, achieving new state-of-the-art performance in Multi-view 3D Object Detection.

\clearpage  

%
%
\bibliographystyle{splncs04}
\bibliography{main}
\end{document}